\pdfoutput=1

\documentclass[11pt]{article}

\usepackage[]{acl}

\usepackage{times}
\usepackage{latexsym}
\usepackage{CJKutf8}

\usepackage[utf8]{inputenc}
\usepackage{graphicx}
\usepackage{inconsolata}

\usepackage[T1]{fontenc}

\usepackage[utf8]{inputenc}

\usepackage{microtype}

\usepackage{amsmath,amsfonts,bm}

\def\eqref#1{equation~\ref{#1}}

\def\1{\bm{1}}

\def\vtheta{{\bm{\theta}}}

\def\vx{{\bm{x}}}
\def\vy{{\bm{y}}}

\DeclareMathAlphabet{\mathsfit}{\encodingdefault}{\sfdefault}{m}{sl}
\SetMathAlphabet{\mathsfit}{bold}{\encodingdefault}{\sfdefault}{bx}{n}

\newcommand{\E}{\mathbb{E}}
\newcommand{\Ls}{\mathcal{L}}

\usepackage{defs}
\usepackage{defs2}
\usepackage{defs3}
\usepackage{clues_defs}
\usepackage{attention}
\usepackage{math-common}
\usepackage{makecell}

\usepackage{caption}
\usepackage{subcaption}
\usepackage{xcolor}
\usepackage{placeins}  %
\usepackage{balance} 
\usepackage{algorithm}
\usepackage{stfloats}

\definecolor{chromeyellow}{rgb}{1.0, 0.65, 0.0}

\newcommand{\nop}[1]{}

\title{Self-Vocabularizing Training for Neural Machine Translation}

\author{
Pin-Jie Lin$^{\spadesuit}$ \, Ernie Chang$^{\blacklozenge}$ \, Yangyang Shi$^{\blacklozenge}$ \, Vikas Chandra$^{\blacklozenge}$ \, \\
$^{\spadesuit}$Virginia Tech \\
$^{\blacklozenge}$Meta \\
\texttt{pinjie@vt.edu, erniecyc@meta.com}
}

\begin{document}
\maketitle

\begin{abstract}

Past vocabulary learning techniques identify relevant vocabulary before training, relying on statistical and entropy-based assumptions that largely neglect the role of model training.
Empirically, we observe that trained translation models are induced to use a byte-pair encoding (BPE) vocabulary subset distinct from the original BPE vocabulary, leading to performance improvements when retrained with the induced vocabulary.
In this paper, we analyze this discrepancy in neural machine translation by examining vocabulary and entropy shifts during self-training—where each iteration generates a labeled dataset by pairing source sentences with the model’s predictions to define a new vocabulary.
Building on these insights, we propose \emph{self-vocabularizing training}, an iterative method that self-selects a smaller, more optimal vocabulary, yielding up to a $1.49$ BLEU improvement.
Moreover, we find that deeper model architectures lead to both an increase in unique token usage and a 6–8\% reduction in vocabulary size.

\end{abstract}

\section{Introduction}

Vocabulary construction, also known as vocabularization, is essential for many natural language processing tasks that involve neural networks, including neural machine translation (MT), as highlighted in various studies~\citep{DBLP:conf/nips/MikolovSCCD13,attention,DBLP:conf/emnlp/GehrmannDR18,DBLP:journals/widm/ZhangWL18,DBLP:conf/naacl/DevlinCLT19}.  
However, past vocabulary learning techniques rely on corpus statistics such as entropy~\cite{xu2020vocabulary} or frequency counts~\cite{DBLP:conf/acl/SennrichHB16a}, without considering contextual information or the model’s ability to represent it.  

\begin{figure}[t] 
    \centering 
    \includegraphics[width=0.85\columnwidth]{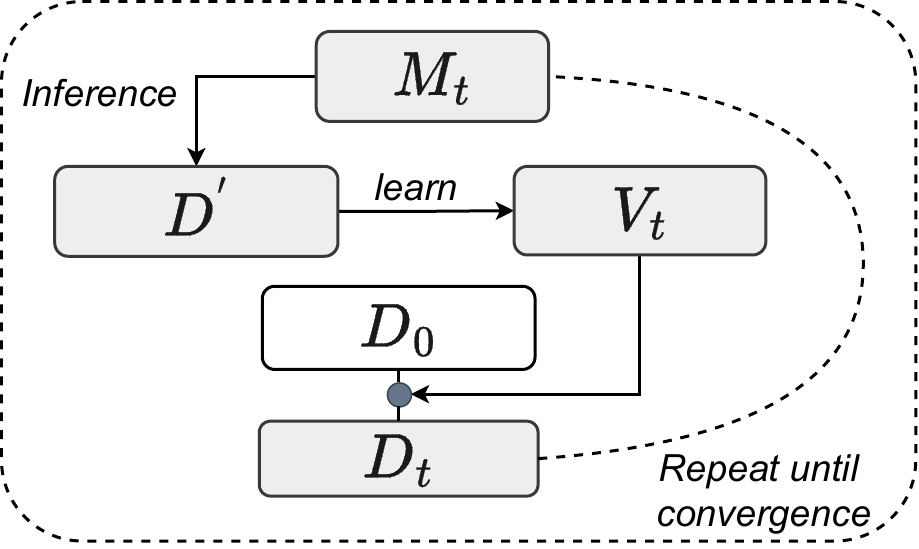} 
    \caption{\small Illustration of self-vocabularizing training: At each iteration, the original dataset $D_0$ is segmented using vocabulary $V_t$ to form the training set $D_t$. $D_t$ is then used to train model $M_t$, which generates a pseudo dataset $D'$. A new vocabulary set $V_{t+1}$ is derived from $D'$, completing the training loop. This process repeats until no further improvements are observed.}
    \label{fig:overview} 
\end{figure}

Despite the success of vocabularization in improving MT model efficiency~\cite{xu2020vocabulary}, we observe a discrepancy between the original byte-pair encoding (BPE) vocabulary~\cite{gage1994new} ($V_0$), derived from the initial training data, and the BPE vocabulary induced from pseudo-labeled data ($V_1$) (see Figure~\ref{fig:overview}).  
This discrepancy is surprising, as it suggests that MT models implicitly learn a pseudo-“optimal” vocabulary ($V_1$) that is substantially different from the original vocabulary ($V_0$) and is also smaller in size.  
For instance, on the IWSLT14 DE-EN dataset, $|V_1|$ is approximately 20\% smaller than $|V_0|$.  
Moreover, MT models retrained with the pseudo vocabulary $V_1$ outperform those trained on the original vocabulary set.  
This suggests a need to re-examine the assumptions underlying vocabulary learning techniques such as byte-pair encoding and the marginal utility of vocabularization~\cite{xu2020vocabulary}, as existing methods may overlook model-data interactions, leaving key optimization factors unaccounted for.  

In this paper, we aim to understand this discrepancy in neural machine translation models by analyzing shifts in vocabulary and entropy during self-training.  
To this end, we conduct experiments on two language tasks, comparing the vocabulary sets learned from the original training data and the pseudo-labeled data.  
Our results suggest limited overlap between the two vocabularies and that pseudo data induces a more optimal vocabulary, enabling further improvements.  
Furthermore, our findings indicate that the decoder has a limited impact on this vocabulary shift, whereas encoder-based interactions play a crucial role in entropy reduction.  
This suggests that future vocabulary induction methods should focus more on the cross-attention module.  
Finally, our study has implications for defining an optimal vocabulary set in language generation.  

This preliminary study also introduces a simple yet effective technique: iterative self-training to self-select a more optimal vocabulary set for performance gains.  

In summary, this paper makes the following contributions:  
(1) We identify a discrepancy between the optimal and pseudo-labeled vocabulary derived from MT models.  
(2) We analyze shifts in vocabulary and entropy during self-training.  
(3) We propose a simple approach to obtain a competitive vocabulary set and introduce a self-vocabularizing training algorithm that improves performance.

\section{Iterative Self-Vocabularization}

Current vocabularization techniques adopt two contrasting perspectives:  
(1) Focusing on frequency or entropy statistics to avoid the computational cost of trial training~\cite{xu2020vocabulary}, which often neglects important parameters and interactions in the process.  
(2) Obtaining a more optimal vocabulary set through training~\cite{DBLP:journals/mt/SaleskyRCNN20}, but at a higher computational cost. 
This work combines both perspectives by adopting an entropy-based vocabularization approach while utilizing self-training.

In self-training, a base model $M_{t}$ is trained on the dataset to generate predictions for input sequences, which are then used to update the next iteration of the base model $M_{t+1}$. 
This process is repeated iteratively with the supervised loss $\Ls$ from labeled instances~\cite{he2019revisiting}, where $x$ and $y$ are the source and target texts, respectively:

\begin{equation}
\label{eq:st-er}
\Ls = -\E_{\vx \sim p(\vx)}\E_{\vy \sim p_{\vtheta^{*}}(\vy|\vx)} \log p_{\vtheta}(\vy|\vx),
\end{equation} 

where $p(\vx)$ is the empirical data distribution approximated with samples from $D$, and $p_{\vtheta}(\vy|\vx)$ is the conditional distribution defined by the model.  
The parameter $\vtheta^{*}$ is randomly initialized at every iteration.  
For each training iteration at $t+1$, we relearn the BPE vocabulary using the \emph{original source} and the \emph{pseudo target} generated with $M_{t}$ (i.e., $D' = \{(\vx, f_{\vtheta}(\vx)) | \vx \in U\}$).  
Then, $M_{t+1}$ is trained on $D_{t+1}$, segmented with the newly derived vocabulary $V_{t+1}$.  

\paragraph{Measuring Vocabulary Shifts}

Subword-based approaches like byte-pair encoding are widely used and have demonstrated strong empirical performance~\citep{DBLP:conf/acl/SennrichHB16a,DBLP:conf/aaai/Al-RfouCCGJ19,DBLP:conf/acl/Costa-JussaF16,DBLP:journals/tacl/LeeCH17,DBLP:conf/mtsummit/DingRD19,deeptransformer,scalingnmt,DBLP:conf/emnlp/KudoR18,DBLP:conf/aaai/WangCG20}.  
These methods construct vocabulary by selecting high-probability subword units.  

Following \citet{xu2020vocabulary}, we define the vocabulary shift as the negative change in entropy normalized by vocabulary size:

\begin{equation}
     \frac{-(\mathcal{H}_{M_{t+1}(\vx)}-\mathcal{H}_{M_{t}(\vx)})}{|V_{t}|},
    \label{eq:id}
\end{equation}

where $M_{t+1}(\vx) \rightarrow V_{t+1}$ and $M_{t}(\vx) \rightarrow V_{t}$ represent vocabularies from two consecutive training iterations, with sizes $|V_{t+1}|$ and $|V_{t}|$.  
The ratio $|V_{t+1}|/|V_{t}|$ reflects compression in vocabulary size.  

Corpus entropy $\mathcal{H}_v$ with vocabulary $V$ is defined as the sum of token entropy, normalized by the average token length:

\begin{equation}
   \mathcal{H}_{v} = - \frac{1}{l_{v}}\sum_{j \in V } p(j)\log p(j),
   \label{eq:entropy}
\end{equation}

where $p(j)$ represents the relative frequency of token $j$ in the training corpus, and $l_{v}$ is the average token length (i.e., the number of characters per token).

\section{Experimental Settings}

\begin{figure*}[!t]
\centering
    \begin{subfigure}{0.48\textwidth}
        \centering
        \includegraphics[width=\textwidth]{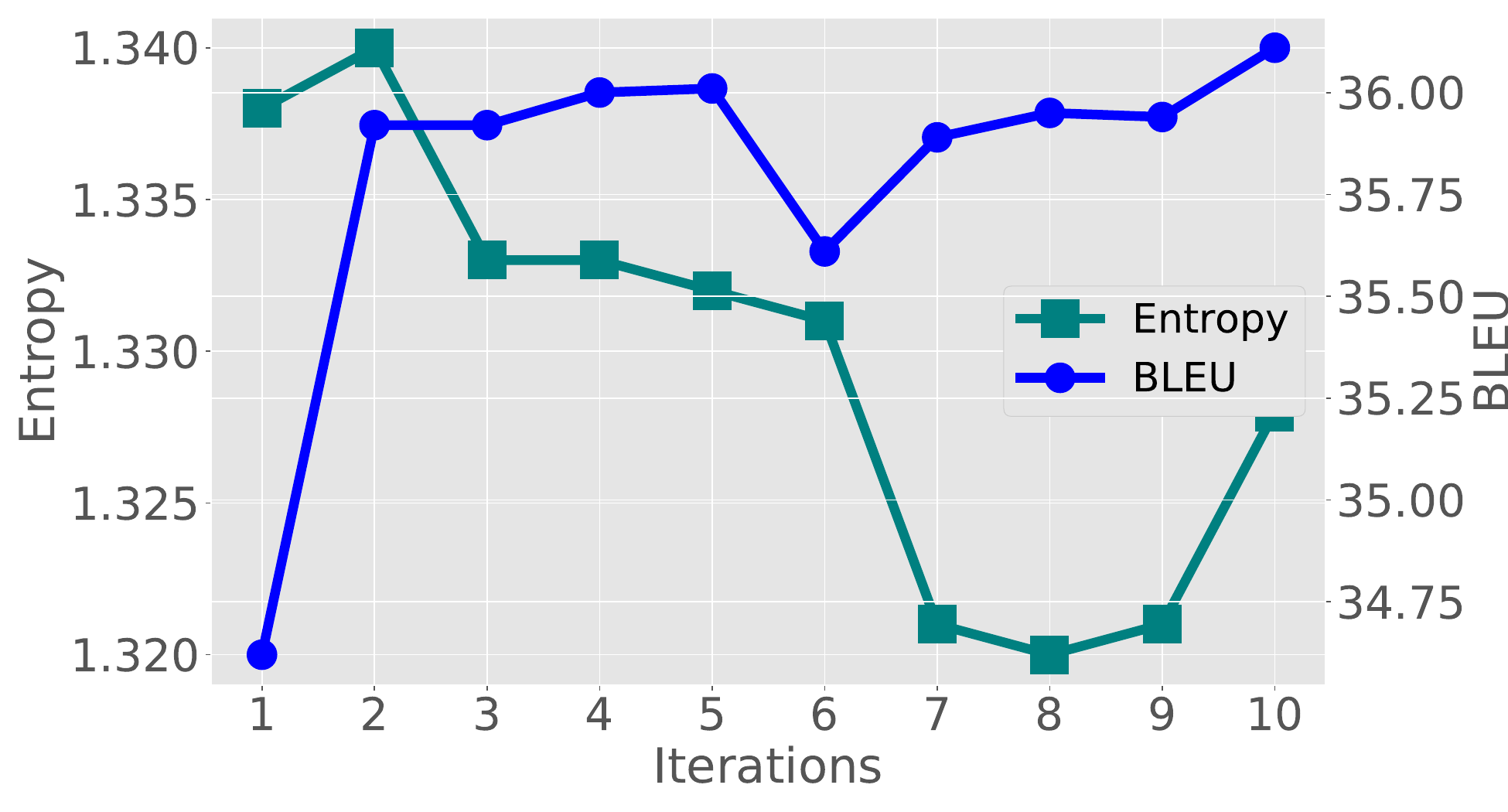}
    \end{subfigure}
    \begin{subfigure}{0.48\textwidth}
        \centering
        \includegraphics[width=\textwidth]{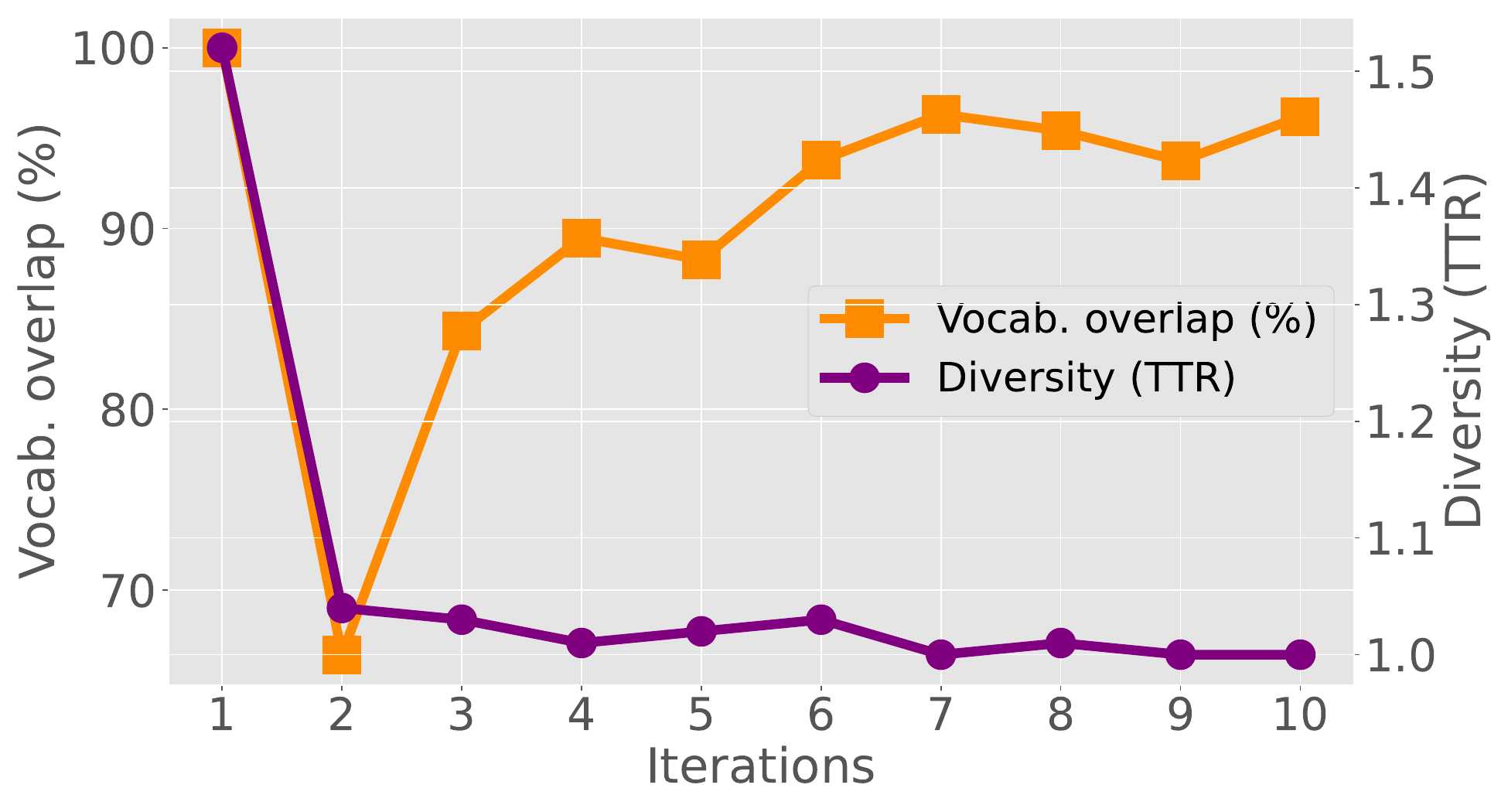}
    \end{subfigure}
    \caption{Entropy and performance across self-vocabularizing training iterations.  
    (\textbf{Left}) BLEU score ({\color{blue}blue}) consistently improves across iterations. Meanwhile, the self-learned vocabulary reduces corpus entropy ({\color{teal}teal}), indicating a better estimation of token distribution.  
    (\textbf{Right}) Vocabulary shift measured by vocabulary overlap ({\color{chromeyellow}orange}) between consecutive vocabularies $V_t$ and $V_{t-1}$, showing that the model initially selects a broad set of subwords before consolidating onto a subset of $V_{t-1}$\footnotemark.  
    The type-token ratio (TTR) ({\color{violet}purple}) reflects the diversity of learned semantic units, reported on the training corpus scaled by $1000$.}
    \label{fig:corpus_shift}
\end{figure*}

For our experiments, we used the IWSLT14 German-English parallel corpus for both German-to-English (DE-EN) and English-to-German (EN-DE) translation tasks. We preprocessed the data using \texttt{MOSES}~\cite{bollmann-etal-2021-moses} and applied byte-pair encoding (BPE)~\cite{DBLP:conf/acl/SennrichHB16a} to construct the vocabulary set. 

We trained a transformer-based NMT model using the \texttt{fairseq} library~\cite{ott-etal-2019-fairseq}, with six layers, four attention heads, and a hidden size of 1024 dimensions. The Adam optimizer was used with a learning rate of 0.0002 and a batch size of 64. Training lasted for 50 epochs, with exponential learning rate decay and early stopping based on the validation set. 

We evaluated model performance using BLEU scores on the test set, comparing against the baseline and other comparable models. For self-training, we ran each iteration until performance converged. In each iteration $i$, the model was trained from scratch in the self-training step (\emph{ST-i}) for analysis purposes. Results were averaged over three initialization runs with different $\vtheta^{*}$.  

\footnotetext[\thefootnote]{Vocabulary overlaps at firs iteration leverages the identical vocabulary where $V_{0}=V_{1}$.}

\subsection{Main Results}

\begin{figure*}[!t]
\centering
    \begin{subfigure}{0.48\textwidth}
        \centering
        \includegraphics[width=\textwidth]{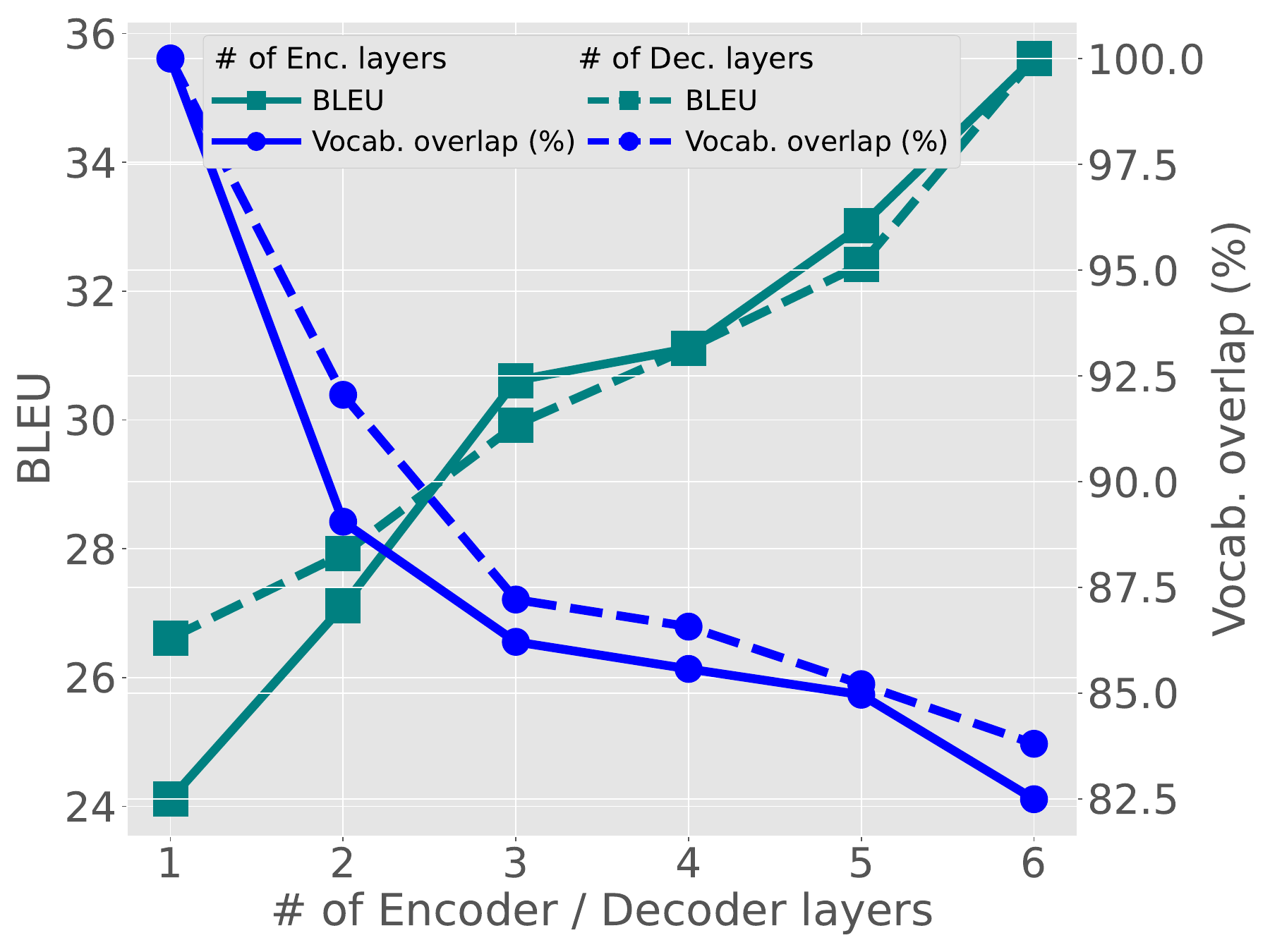}
    \end{subfigure}%
    \begin{subfigure}{0.48\textwidth}
        \centering
        \includegraphics[width=\textwidth]{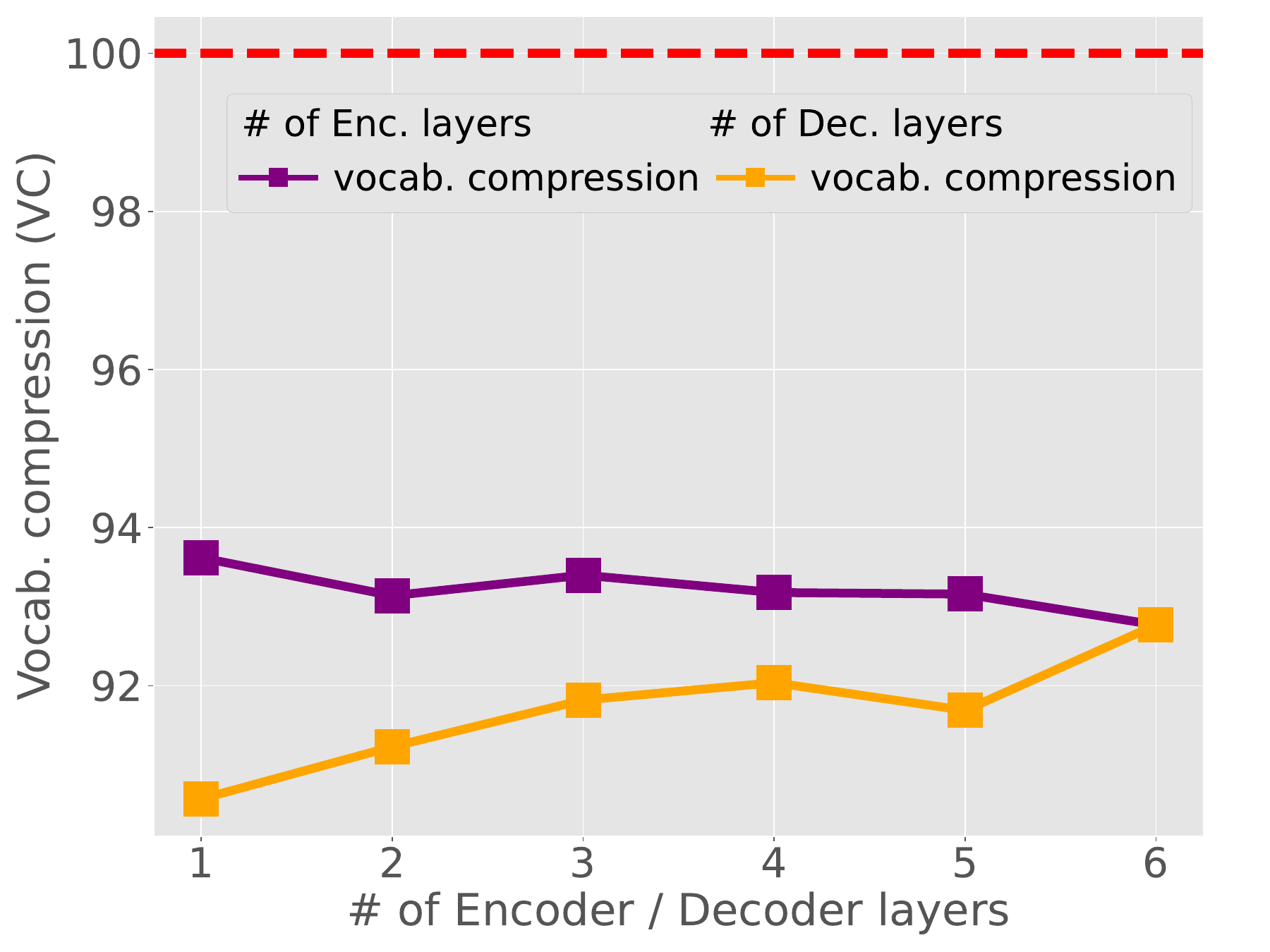}
    \end{subfigure}%
    \caption{\small 
    Performance and vocabulary overlap across models with different encoder and decoder depths.  
    (\textbf{Left}) As the number of encoder ({\color{teal}teal -}) or decoder ({\color{teal}teal - -}) layers increases, BLEU scores consistently improve. However, vocabulary overlap decreases for deeper encoder ({\color{blue}blue -}) or decoder ({\color{blue}blue - -}) layers, indicating that deeper models tend to use more unique tokens.  
    (\textbf{Right}) Vocabulary compression (VC) across models with varying depths. All models trained with self-vocabularizing training effectively compress the token set. Notably, deeper encoder models ({\color{purple}purple}) exhibit a smoother reduction in VC rates, whereas deeper decoder models ({\color{orange}orange}) require more tokens for inference. VC is reported on the test set using models of different depths in either the encoder or decoder, with a single round of self-vocabularizing training.}
    \label{fig:vocab_shift_iterations}
\end{figure*}

We first compared the performance of MT models trained with two different approaches: one using a fixed output vocabulary and the other refining the output vocabulary through self-training iterations.  
As shown in Table~\ref{table:main} and Figure~\ref{fig:vocab_shift_iterations}, the MT model trained with the self-trained output vocabulary gradually improves with each newly derived vocabulary, achieving up to a $1.3$ BLEU point increase after a single iteration.  
Table~\ref{table:main} further confirms a consistent trend across both language tasks: self-training improves model performance and reduces vocabulary entropy, leading to enhanced fluency and correctness while decreasing vocabulary size.

\begin{table}[ht]
\centering
\resizebox{\columnwidth}{!}{
\begin{tabular}{c|c|c|c|c|c}

& \textbf{BLEU} & \textbf{|V|} & \textbf{Overlap (\%)} &  \textbf{Fluency} & \textbf{Adequacy} \\ \hline
\textbf{ST-0} & 34.62 & 10000  & -            & 2.89   & 3.21     \\ \hline
\textbf{ST-1} & 35.92 & 8950 & 66.42           &  3.13   & 3.46     \\ \hline
\textbf{ST-5} & 36.01 & 8892 & 88.27          & 3.42   & 3.87      \\ \hline
\textbf{ST-10} & \textbf{36.11} & 8702 & 96.19            & 3.95    & 4.21     \\ 
\end{tabular}}
\caption{\small Performance comparison of BLEU, vocabulary size ($|V|$), vocabulary overlap (\%), fluency, and adequacy on the IWSLT14 DE-EN translation task for \textbf{ST-0}, \textbf{ST-1}, \textbf{ST-5}, and \textbf{ST-10} models.  
Fluency and adequacy scores are segment-level averages on 100 random outputs, rated on a 1-5 scale (5 being the most fluent or correct)~\cite{koehn-monz-2006-manual,freitag-etal-2021-experts}. Scores were assigned by three raters and then averaged.  
Detailed results for IWSLT14 EN-DE are provided in Appendix \ref{app_iwslt14_en_de}.}\label{table:main}
\end{table}

Beyond translation quality, we also observe lower overall corpus entropy and a smaller vocabulary in the self-trained model (see Figure~\ref{fig:corpus_shift}).  
This suggests that self-training not only enhances translation accuracy but also results in a more efficient model with a compact, more targeted vocabulary—potentially enabling faster and more memory-efficient deployment.

\section{Ablations of Self-Vocabularization}

\subsection{Shifts Across Iterations}

In text generation, self-training can enhance the quality of the generated output.  
However, the impact of the number of self-training iterations on output entropy (i.e., the randomness or unpredictability of the generated text) is not straightforward and depends on the specifics of the model and training data.  
We therefore examine:  
(1) corpus entropy and  
(2) subword-based overlap between the original and self-trained BPE vocabulary.  

\paragraph{Corpus Entropy.}  

Increasing the number of self-training iterations allows the model to learn from a progressively smaller set of labeled examples, potentially leading to more coherent and accurate outputs with reduced diversity.  
In the left plot of Figure~\ref{fig:vocab_shift_iterations}, we observe that, in general, as entropy gradually decreases, self-training performance improves until the rate of change in both entropy and BLEU slows.  
Surprisingly, even at the $10^{th}$ iteration, the model continues to improve its BLEU score.  

\paragraph{Vocabulary Overlap ($V_{O} \cap V_{P}$).}  

The original BPE vocabulary consists of subword units created by applying BPE to the training data, serving as a fixed vocabulary during training and inference.  
While the number of self-training iterations does not directly alter the BPE vocabulary (as it is predefined before training), fine-tuning on additional labeled examples can improve model performance, leading to more accurate and diverse outputs that better align with the original BPE vocabulary.  
Additionally, vocabulary size consistently decreases across iterations.  
We observe an initial sharp drop of approximately 10\% after the first iteration, followed by a gradual reduction in BPE vocabulary size until the $5^{th}$ iteration (see Figure~\ref{fig:corpus_shift}).  

\subsection{Ablations on Model Architecture}

\paragraph{Deeper Model Depth Contributes to Lower Vocabulary Overlap.}  

The number of encoder and decoder layers in a neural network plays a crucial role in determining output coherence and accuracy, which in turn affects the model’s output token set.  
As shown in Figure~\ref{fig:vocab_shift_iterations}, increasing encoder or decoder layers generally improves BLEU scores.  
Additionally, vocabulary overlap gradually decreases to approximately 93\% as the number of layers increases, following a similar trend observed in Figure~\ref{fig:corpus_shift} at the first iteration.  
This suggests that deeper architectures allow the model to implicitly select more unique tokens compared to shallow models, with encoders playing a particularly important role in vocabulary selection.  

\paragraph{Vocabulary Compression.}  

\emph{Vocabulary compression (VC)} is defined as the ratio of the number of tokens used in the inference output to the number of tokens in the original test set, i.e., $\frac{|V|^{inf.}}{|V|^{test}}$
Figure~\ref{fig:vocab_shift_iterations} illustrates the relationship between the number of encoder/decoder layers and VC.  
All models achieve significant token set compression, reducing vocabulary size by 6\% to 8\%.  
Notably, increasing encoder depth results in a smaller token set, whereas increasing decoder depth leads to a larger token set.  
We conjecture that deeper encoders have a stronger ability to process source sentences and represent them as fixed-length context vectors, enabling the decoder to use fewer subword units for translation.

\section{Conclusions and Findings}

In this paper, we investigated the discrepancy between the "optimal" vocabulary set identified prior to training a translation model and the vocabulary actually used by the trained model.  
We found that the trained model diverged from the original BPE vocabulary and that a single iteration of self-training was sufficient to generate a competitive vocabulary set.  
Additionally, we examined the relationship between the self-vocabularizing process and the encoder-decoder architecture, demonstrating that deeper models favor the selection of rarer tokens while reducing vocabulary size, whereas decoders have a lesser influence on vocabularization.

\section*{Limitations}

While self-vocabularizing training is simple and provides significant improvements over baseline training, it remains time-consuming.  
Moreover, further analysis is needed to better understand vocabulary shifts and how to efficiently determine the optimal set without requiring costly training iterations.  
This analysis should include an examination of token types and subword granularity, such as how subword segmentation evolves across training iterations.  

In addition, our findings have yet to be verified across multiple language pairs, leaving this as an avenue for future work.  
Overall, this study highlights the need to incorporate vocabulary relearning during self-training and suggests that new vocabulary construction techniques could bridge the gap between model training and text interactions.  

\section*{Ethics Statement}

Vocabularization with model training has significantly improved machine translation performance.  
To minimize potential negative impacts, we conduct our experiments on publicly available datasets commonly used in machine translation research.  
However, if this method is applied to sensitive data, such as medical records, privacy-preserving policies should be strictly considered.  

Additionally, while deeper model architectures promote the use of unique tokens, they also increase computational demands.  
The potential environmental impact of large-scale model training should be carefully evaluated when scaling this approach.

\bibliography{anthology,sample_size,datasets,custom,self_train,volt}
\bibliographystyle{acl_natbib}

\clearpage

\appendix

\section{Detailed Results}

\begin{figure*}[!t]
\centering
    \begin{subfigure}{0.48\textwidth}
        \centering
        \includegraphics[width=\textwidth]{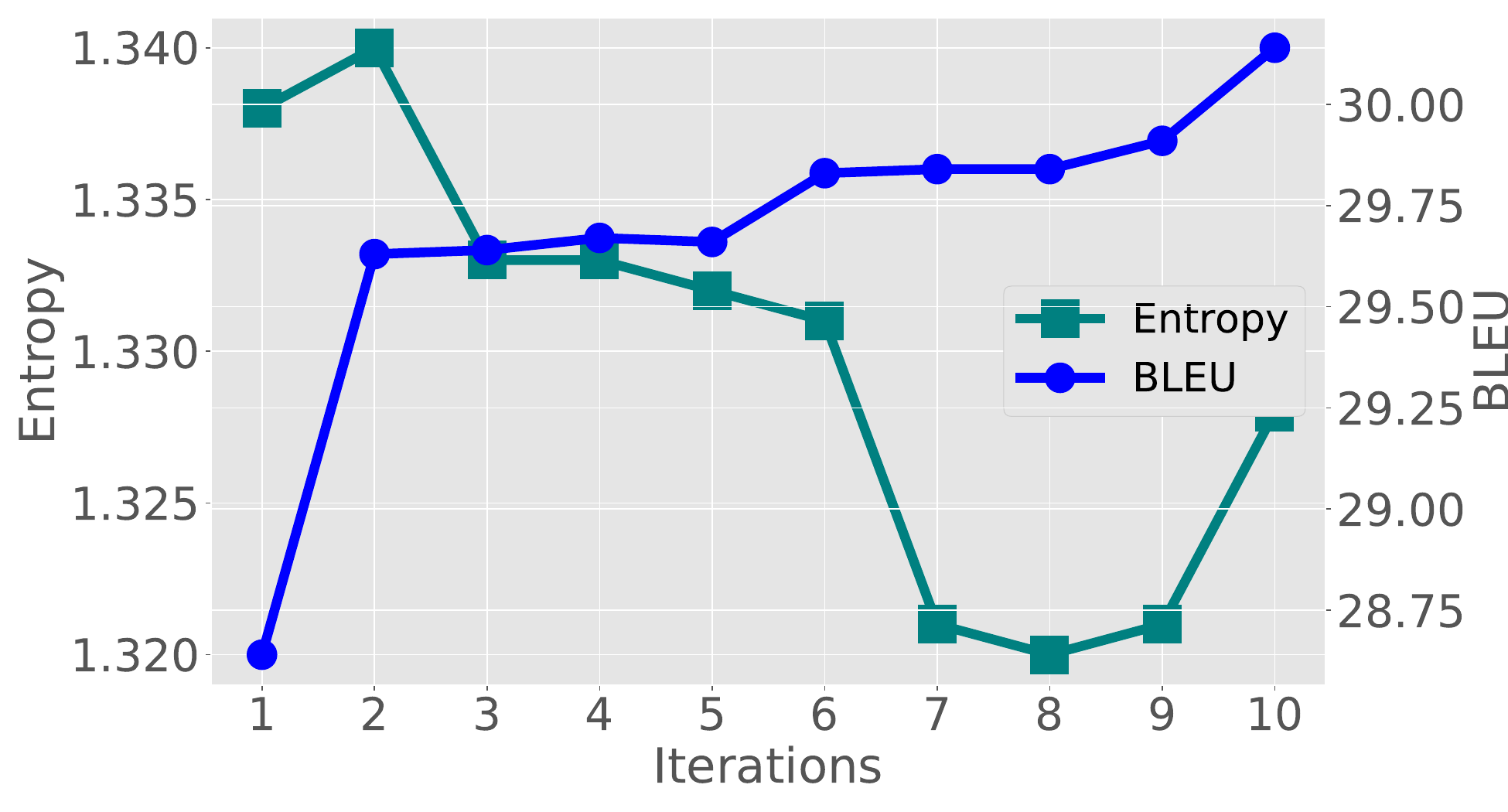}
        \label{fig:corpus_shift_iwslt14_reverse_one}
    \end{subfigure}%
    \begin{subfigure}{0.48\textwidth}
        \centering
        \includegraphics[width=\textwidth]{diagrams/figures/2b.pdf}
        \label{fig:corpus_shift_iwslt14_reverse_two}
    \end{subfigure}%
    \caption{\small 
Impact of self-vocabularizing training on IWSLT14 EN-DE.  
(\textbf{Left}) BLEU scores improve consistently across iterations, while corpus entropy decreases, indicating more stable and predictable token distributions.  
(\textbf{Right}) Vocabulary overlap reduces as the model gradually refines its subword selection, while the type-token ratio (TTR) reflects evolving semantic diversity.}

    \label{fig:stacked_corpus_shift}
\end{figure*}

\subsection{IWSLT14 EN-DE}\label{app_iwslt14_en_de}

We present results for MT models trained with self-vocabularizing on the IWSLT14 EN-DE dataset over 10 iterations.  
We report performance for the baseline model (\textbf{ST-0}) and models trained with self-vocabularizing at iterations $1$, $5$, and $10$.  
As shown in Table~\ref{table:main_iwslt14_reverse}, all self-trained models outperform the fixed-vocabulary baseline (\textbf{ST-0}), with a 2.4-point increase in BLEU after 10 iterations.  
Additionally, vocabulary size decreases with each iteration, leading to a more compact vocabulary.  

\begin{table}[h]
\centering
\resizebox{0.8\columnwidth}{!}{
\begin{tabular}{c|c|c|c}
& \textbf{BLEU} & \textbf{|V|} & \textbf{Overlap (\%)}  \\ \hline
\textbf{ST-0}  & 28.64  & 10000 & -      \\ \hline
\textbf{ST-1}  & 29.63  & 8969  & 66.42  \\ \hline
\textbf{ST-5}  & 29.66  & 8892  & 88.27  \\ \hline
\textbf{ST-10} & 30.14  & 8863  & 93.91  \\ 
\end{tabular}}
\caption{\small BLEU scores, vocabulary size ($|V|$), and overlap (\%) on the IWSLT14 EN-DE translation task for \textbf{ST-0}, \textbf{ST-1}, \textbf{ST-5}, and \textbf{ST-10} models.}
\label{table:main_iwslt14_reverse}
\end{table}

Figure~\ref{fig:stacked_corpus_shift} illustrates the impact of self-vocabularizing training on corpus entropy, performance, vocabulary overlap, and diversity for the IWSLT14 EN-DE dataset.  
We observe a consistent decrease in corpus entropy and vocabulary overlap, alongside performance improvements with increasing training iterations.  
This confirms the effectiveness of self-vocabularizing training on the EN-DE translation task.  
Notably, the EN-DE translation exhibits lower diversity than the DE-EN task, which aligns with expectations since German shares more semantic units than English.

\end{document}